\documentclass[letterpaper, 10 pt, conference]{ieeeconf}  

\usepackage{graphicx}
\usepackage{overpic} 
\usepackage{algorithm}
\usepackage{amsmath}
\usepackage{xcolor}
\usepackage{multirow}
\usepackage{booktabs}
\usepackage[table]{xcolor}
\usepackage{colortbl}
\usepackage{array}
\usepackage{hyperref}
\usepackage[table]{xcolor}   
\usepackage{pgf}               
\usepackage{etoolbox}          

\definecolor{green1}{RGB}{120, 198, 121} 
\definecolor{green2}{RGB}{173, 221, 142} 
\definecolor{green3}{RGB}{217, 240, 199} 
\definecolor{yellow1}{RGB}{255, 237, 160} 
\definecolor{yellow2}{RGB}{254, 217, 118} 
\definecolor{yellow3}{RGB}{254, 196, 79}  
\definecolor{orange1}{RGB}{253, 174, 107} 
\definecolor{orange2}{RGB}{253, 141, 90}  
\definecolor{red1}{RGB}{252, 108, 96}     
\definecolor{red2}{RGB}{227, 74, 74}      

\newcommand{\getcolor}[1]{%
    \ifdim #1pt > 0.95 pt \cellcolor{green1}%
    \else\ifdim #1pt > 0.9pt \cellcolor{green2}%
    \else\ifdim #1pt > 0.85pt \cellcolor{green3}%
    \else\ifdim #1pt > 0.8pt \cellcolor{yellow1}%
    \else\ifdim #1pt > 0.75pt \cellcolor{yellow2}%
    \else\ifdim #1pt > 0.7pt \cellcolor{yellow3}%
    \else\ifdim #1pt > 0.6pt \cellcolor{orange1}%
    \else\ifdim #1pt > 0.5pt \cellcolor{orange2}%
    \else\ifdim #1pt > 0.4pt \cellcolor{red1}%
    \else \cellcolor{red2}%
    \fi\fi\fi\fi\fi\fi\fi\fi\fi%
}

\usepackage[noend]{algpseudocode}
\usepackage{cite}  

\IEEEoverridecommandlockouts                              

\title{\LARGE \bf
FSR-VLN: Fast and Slow Reasoning for Vision-Language Navigation with Hierarchical Multi-modal Scene Graph }

\author{
Xiaolin Zhou$^{1}$$^{*}$, Tingyang Xiao$^{1}$$^{*}$, Liu Liu$^{1}$, Yucheng Wang$^{1}$, Maiyue Chen$^{1}$,\\
Xinrui Meng$^{2}$, Xinjie Wang$^{1}$, Wei Feng$^{1}$, Wei Sui$^{2}$, and Zhizhong Su$^{1}$
\thanks{$^{1}$Horizon Robotics, Beijing, China}%
\thanks{$^{2}$D-Robotics Robotics, Beijing, China}%
\thanks{$^{*}$Equal contribution}%
}

\begin{document}

\maketitle
\thispagestyle{empty}
\pagestyle{empty}


\begin{abstract}
Visual-Language Navigation (VLN) is a fundamental challenge in robotic systems, with broad applications for the deployment of embodied agents in real-world environments. Despite recent advances, existing approaches are limited in long-range spatial reasoning, often exhibiting low success rates and high inference latency, particularly in long-range navigation tasks.
To address these limitations, we propose FSR-VLN, a vision-language navigation system that combines a Hierarchical Multi-modal Scene Graph (HMSG) with Fast-to-Slow Navigation Reasoning (FSR). The HMSG provides a multi-modal map representation supporting progressive retrieval, from coarse room-level localization to fine-grained goal view and object identification. Building on HMSG, FSR first performs fast matching to efficiently select candidate rooms, views, and objects, then applies VLM-driven refinement for final goal selection. 
We evaluated FSR-VLN across four comprehensive indoor datasets collected by humanoid robots, utilizing 87 instructions that encompass a diverse range of object categories. FSR-VLN achieves state-of-the-art (SOTA) performance in all datasets, measured by the retrieval success rate (RSR), while reducing the response time by 82\% compared to VLM-based methods on tour videos by activating slow reasoning only when fast intuition fails.
Furthermore, we integrate FSR-VLN with speech interaction, planning, and control modules on a Unitree-G1 humanoid robot, enabling natural language interaction and real-time navigation. The code and video can be available at \href{https://horizonrobotics.github.io/robot_lab/fsr-vln/}{https://horizonrobotics.github.io/robotlab/fsr-vln/}

\end{abstract}

\begin{figure*}[h]
    \centering
    \vspace{-6pt}
    \includegraphics[width=1.0\linewidth]{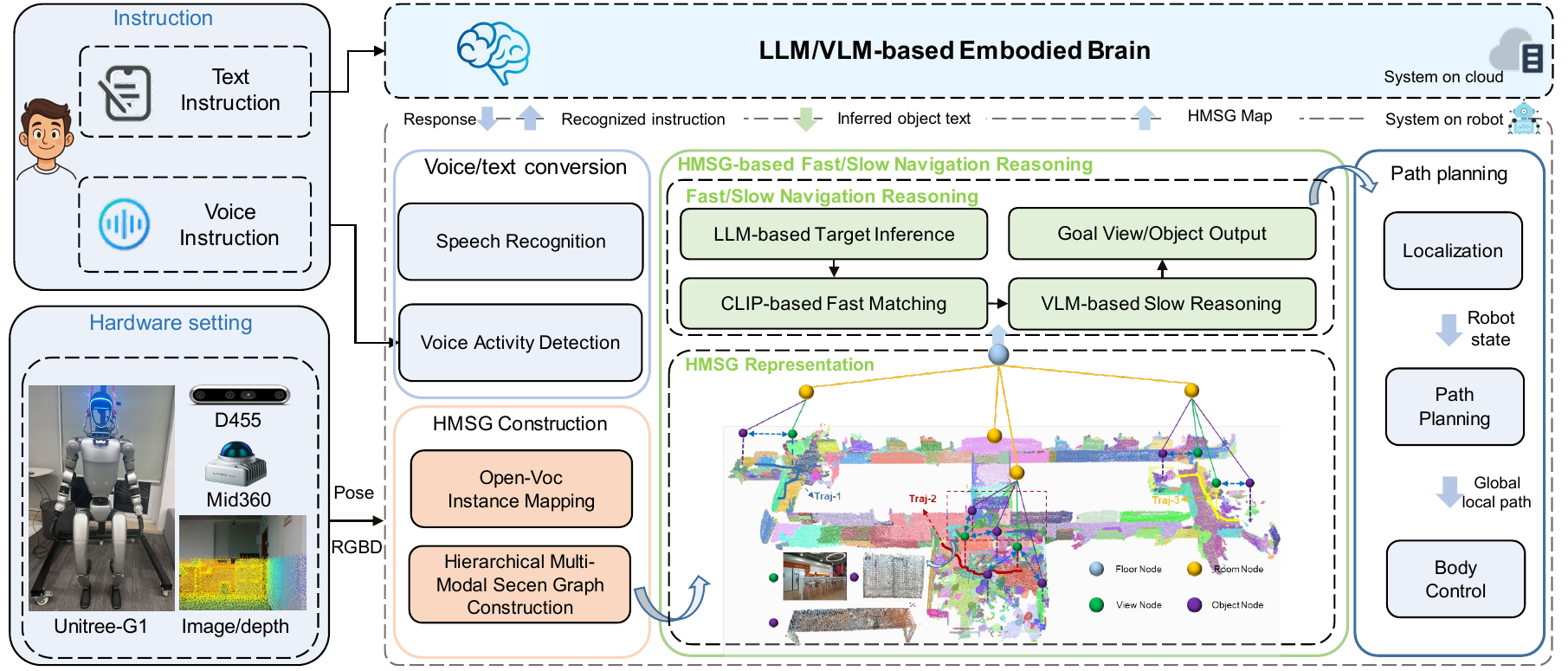}
    \caption{System Overview. The proposed humanoid robotics navigation system integrates HMSG with FSR to achieve view/object-level real-world long-range navigation. Specifically, RGBD and pose data are first utilized to construct HMSG, which provides a hierarchical and multimodal feature-based representation of the environment. During online interaction, the user’s text or voice input is converted into instructions via voice activity detection and speech recognition, and the LLM infers the target object. Based on the HMSG, fast-matching and slow VLM reasoning jointly identify the optimal goal view/object. The identified goals are subsequently used by the global path planning.}
    \label{fig:system_overview}
    \vspace{-7pt}
\end{figure*}

\section{INTRODUCTION}

Visual-Language Navigation (VLN) is a fundamental task in embodied AI, enabling robots to operate effectively in complex real-world environments \cite{8578485}. Despite significant progress in map-free VLN research and growing efforts to equip robots with visual-language reasoning \cite{yokoyama2024vlfm, zhang2024visionandlanguagenavigationtodaytomorrow, long2024instructnavzeroshotgenericinstruction, zhou2023navgptexplicitreasoningvisionandlanguage}, existing methods remain limited in long-range spatial cognition \cite{ramakrishnan2024does}, particularly for long-range navigation. A key bottleneck is the lack of persistent long-range spatial memory, which encodes, organizes, and retrieves environmental knowledge. Such memory allows robots to capture comprehensive spatial relationships and adapt to complex indoor environments \cite{ruan2025reactivecognitivebraininspiredspatial, Gu_2022}. Without it, robots struggle to understand and reason about long-range spaces.
To address this limitation, recent spatial memory-based approaches have explored embedding 2D and 3D geometric maps with semantic features \cite{Peng2023OpenScene,jin2025openfusionopenvocabularyrealtimescene,conceptfusion, FM-Fusion}, integrating dense geometric reconstructions with the pre-trained zero-shot Vision Foundation Model (VFM) such as CLIP \cite{radford2021learningtransferablevisualmodels}, OWL-ViT \cite{minderer2022simpleopenvocabularyobjectdetection}, and LSeg \cite{li2022languagedriven}. These approaches support open-vocabulary object-level retrieval while maintaining high geometric fidelity. Building on geometric semantic maps, methods such as Clio \cite{Maggio2024Clio}, ConceptGraph \cite{gu2023conceptgraphsopenvocabulary3dscene,jiang2024roboexpactionconditionedscenegraph,tang2025openinopenvocabularyinstanceorientednavigation}, RoboExp \cite{jiang2024roboexpactionconditionedscenegraph}, and OpenIN \cite{jin2025openfusionopenvocabularyrealtimescene} leverage open-vocabulary 3D scene graphs to represent long-range environments and provide semantic interfaces for prompting large language model (LLM). Parallel work, including HOVSG \cite{werby23hovsg} and IRS \cite{chen2025irsinstancelevel3dscene}, abstracts these dense maps into hierarchical structures spanning floors, rooms, and objects, supporting efficient semantic object retrieval for long-range navigation.

Although 2D and 3D semantic maps and 3D scene graphs provide geometry-consistent and hierarchical spatial memory, they rely on pre-extracted visual features\cite {zhang2024visionandlanguagenavigationtodaytomorrow,pekkanen2025visuallanguagegridmapscapture}, lack direct interaction with a vision-language model (VLM), and are sensitive to geometric noise. These limitations, in turn, hinder their adaptability to diverse user instructions and complex real-world navigation \cite{huang23vlmaps,raychaudhuri2025semanticmappingindoorembodied}.

Inspired by human navigation, where individuals reason over previously observed images or objects without detailed 3D maps, image-based topological navigation has emerged as a promising alternative \cite{savinov2018semiparametric, shah2023gnmgeneralnavigationmodel, xu2024mobility, 10610234}. By retaining raw image data, image-based topological maps facilitate interaction with LLM and VLM, such as GPT-4o and Gemini, which excel at reasoning over both visual and textual inputs \cite{li2025survey}. Despite their high success rate in image-level navigation, existing methods lack explicit 3D geometric structure and usually rely on video captioning, which is inefficient for reasoning over long sequences \cite{anwar2024remembrbuildingreasoninglonghorizon}. 

To address these challenges, we introduce \textbf{FSR-VLN}, a novel \textbf{F}ast and \textbf{S}low \textbf{R}easoning system for vision-language navigation. The contributions of our paper are as follows.
\begin{enumerate}
    \item  To the best of our knowledge, we are the first to introduce \textbf{Hierarchical Multi-modal Scene Graph (HMSG)} map representation. Our approach elegantly synthesizes the strengths of hierarchical scene graphs for long-range navigation with image topological graphs for fine-grained reasoning, which is essential for LLM/VLM-based systems. This unique hybrid representation facilitates progressive retrieval, from coarse-grained navigational cues to precise goal localization, all while preserving multi-modal information to ensure robustness and high success rates in real-world environments.
    \item  Inspired by the dual-process theory of human cognition, we introduce a novel \textbf{Fast-to-Slow Navigation Reasoning} system. The system operates in two distinct stages: a fast, intuitive matching step retrieves candidate views and objects, followed by a slow, deliberate reasoning stage where an VLM verifies and refines the final goal. This multi-stage architecture seamlessly integrates efficient feature-space matching with robust VLM-driven visual verification, leading to a significant improvement in success rates for real-world, long-range navigation.
    \item Evaluated on 87 robot-collected instructions across four diverse categories in long-range real-world indoor environments, FSR-VLN achieves state-of-the-art (SOTA). It achieves 77\% higher success rates than HOVSG and 167\% higher than MobilityVLA through superior object localization, while reducing average response time by 82\% compared to MobilityVLA.
    
    \item By integrating FSR-VLN with speech interaction, planning, and control modules on the Unitree-G1 humanoid robot, we present a comprehensive humanoid robotics navigation system that is capable of operation guided by natural language.
\end{enumerate}

\section{Related Works}

\subsection{Object Navigation with Geometric Semantic Map  }

Recent works have extended geometric point cloud maps by incorporating semantic features to enable object navigation (ObjNav). Systems such as OK-Robot \cite{Liu_2024}, VLMap \cite{huang23vlmaps}, OVL-MAP \cite{10879413}, and BeliefMapNav \cite{zhou2025beliefmapnav3dvoxelbasedbelief} embed features from Vision Foundation Models (VFMs, e.g., CLIP \cite{radford2021learningtransferablevisualmodels}, OWL-ViT \cite{minderer2022simpleopenvocabularyobjectdetection}, LSeg \cite{li2022languagedriven}) into voxel grids. While this approach preserves geometric consistency and supports open-vocabulary queries, it suffers from several critical limitations. Despite these advances, the performance of the systems remains limited by the inherent weaknesses of VFMs and their sensitivity to 3D reconstruction noise and odometry drift. Reliance on a single map representation further restricts integration with LLMs and VLMs, which excel at cross-modal reasoning and generalization.

Building on semantic point cloud maps, scene graph-based methods represent objects and spatial concepts as nodes, with their relations encoded as edges, providing compact and expressive models of long-range environments \cite{armeni20193d,rosinol20203d,hughes2022hydra}. This object-centric decomposition supports higher-level reasoning for navigation and manipulation \cite{werby23hovsg}. While many works employ 3D scene graphs \cite{rosinol20203d,hughes2022hydra,greve2024collaborative} to efficiently model large environments, most rely on closed-set semantics. To enable open-vocabulary queries, approaches such as ConceptGraphs \cite{gu2024conceptgraphs}, HOVSG \cite{werby23hovsg}, DOVSG \cite{yan2025dynamicopenvocabulary3dscene}, and OpenIN \cite{tang2025openin} construct open-vocabulary 3D scene graphs, which can also interface with LLM.

However, these approaches typically depend on rigid instruction formats (e.g., “object A in region B on floor C”) and semantic similarity–based retrieval, which limit flexibility and hinder full exploitation of VLM reasoning. In contrast, our method leverages LLM and VLM to interpret diverse user instructions and to refine semantic retrieval, thereby improving navigation success in real-world environments.

\subsection{ObjNav with Image-based Topological Graph}

Several recent works represent environments as image-based topological graphs to facilitate VLN. MobilityVLA \cite{xu2024mobility} introduces a vision-language action (VLA) framework that leverages long-context VLM for goal frame retrieval and uses topological graphs for waypoint planning. Uni-NaVid \cite{zhang2024uni} adopts a video-based VLA approach, retrieving the most visually similar image to a query object and generating robot actions in an end-to-end manner. ReMEmbR \cite{anwar2025remembr} emphasizes spatio-temporal memory by storing caption embeddings with pose and time metadata for retrieval-augmented reasoning. More recently, MapGPT \cite{chen2024mapgpt} incorporates map-guided prompting and adaptive path planning, significantly improving zero-shot generalization. Astra \cite{chen2025astra} extends image-based topological maps with landmark information, while RoboHop \cite{garg2024robohop} employs a zero-shot “segment servoing” strategy to reach object subgoals. Building on RoboHop’s open-set navigation pipeline, TANGO \cite{podgorski2025tango} generates sub-object goals through a global path planner grounded in an object-level topological graph.

Although these methods preserve rich semantic cues and achieve strong performance in image-level reasoning, they rely solely on video captioning, which is inefficient in long video contexts and often leads to mismatches between semantic and geometric information \cite{chen2025astra}. In real-world environments, however, abundant spatial information-such as room numbers, equipment labels, and other annotations can significantly improve navigation performance \cite{anwar2025remembr}. To leverage this, we propose a hierarchical multi-modal scene graph representation that encodes both spatial and semantic features across floors, rooms, views, and objects. Building on these multi-modal features, we introduce a fast-to-slow reasoning mechanism to enhance navigation success rates.

\section{Method}

Our primary contribution is centered on map representation and navigation reasoning. To demonstrate the practical utility of our FSR-VLN system, we integrate it with speech interaction, planning, and control modules on the Unitree-G1 humanoid robot, forming a complete real-world navigation system.

As illustrated in Fig.~\ref{fig:system_overview}, we first build HMSG with real-world LiDAR–camera datasets collected by Unitree-G1. Specifically, the SOTA SLAM system FAST-LIVO2 \cite{10757429} is used to extract RGBD data and poses to create an instance-level open-vocabulary map. Building on an instance-level open-vocabulary map \cite{werby23hovsg}, we construct the Hierarchical Multi-modal Scene Graph (HMSG), which divides the environment into four levels-floor, room, view, and object. Each node is enriched with multi-modal features, including geometric attributes, semantic information, and topological connections.

During online interaction, the speech recognition module \cite{gao2023funasr} converts user speech into text, which is then parsed by an LLM to extract the corresponding query. In the FSR pipeline, the query text is first grounded to candidate regions and objects through CLIP-based similarity matching and then refined via VLM-based visual verification to ensure robust and accurate navigation. By combining the HMSG representation with fast-to-slow reasoning (FSR), FSR-VLN enables efficient and interpretable navigation, particularly in long-range and cluttered real-world environments. Further details of this mechanism are provided in Section~\ref{Coarse-to-fine navigation reasoning}. After obtaining the target object’s coordinates, the system performs path planning and whole-body control to reach the goal. 

\subsection{Hierarchical Multi-modal Scene Graph Representation} 
\begin{figure}[ht]
    \vspace{-5pt}
    \centering
    \includegraphics[width=1.0\linewidth]{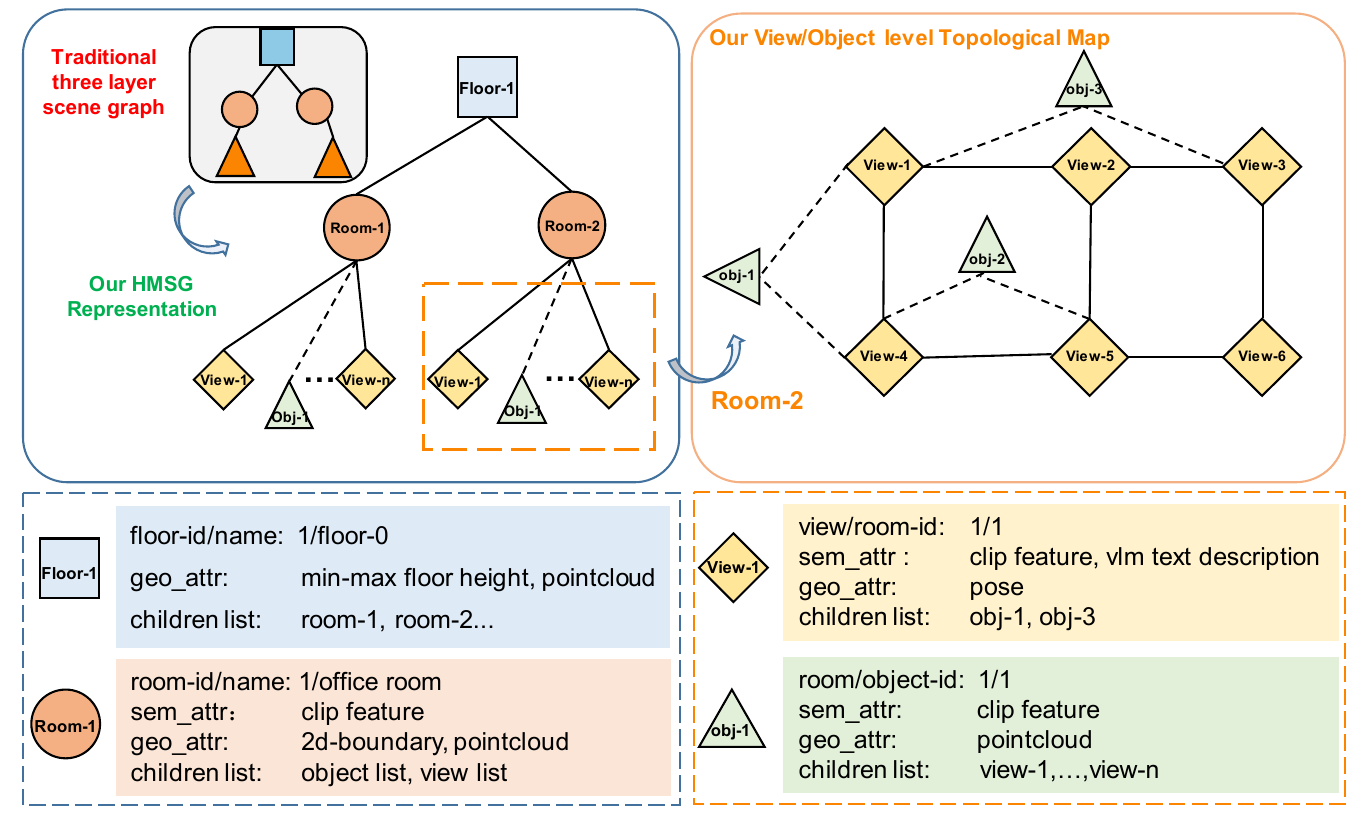}
    \caption{HMSG representation. Our proposed HMSG is a four-level hierarchy: floor, room, view, and object nodes. Each node contains multi-modal features, including geometric attributes, semantic attributes, and topological relationships.}
    \label{fig:SceneGraph}
    \vspace{-6pt}
\end{figure}
As shown in Fig.\ref{fig:SceneGraph}, the proposed HMSG is organized into four levels: floor, room, view, and object nodes. Each node encodes multi-modal features, including geometric attributes, semantic attributes, and explicit topological relationships. This design supports layer-wise retrieval and enables fast-to-slow navigation reasoning with LLM and VLM.

\textbf{Floor nodes} record unique floor identifiers, names, geometric attributes (e.g., min/max heights, PLY point clouds), and references to contained room nodes.

\textbf{Room nodes} store ID, 2D polygon boundaries, point clouds, semantic attributes (name, CLIP embeddings), and links to associated view and object nodes. They support long-range object-level navigation across rooms.

\textbf{Object nodes} represent discrete instances within a room. Each node has geometric properties (3D bounding boxes, point clouds), semantic embeddings, and links to parent room nodes and image views.

Prior scene graph methods typically use floor, room, and object nodes, relying solely on CLIP matching, which constrains visual–spatial understanding. While image-based topological graphs with VLM reasoning improve navigation success, they lack 3D structure and incur high computational cost. To address these limitations, we introduce \textbf{view nodes} as a dedicated layer to capture spatial and perceptual relationships between image views and objects. This layer enables reasoning over image views with VLM, allowing the robot to select contextually relevant views for image-level navigation while enhancing object-level localization.

\textbf{View nodes} represent specific visual perspectives within a room. Each node stores CLIP embeddings and VLM-generated descriptions as semantic features, and camera poses as geometric attributes. They are linked to visible object nodes, encoding visibility relationships that support multi-view perception and image- or object-level grounding. Undirected edges between view nodes are defined by relative poses, enabling global path planning.

\subsection{Hierarchical Multi-modal Scene Graph Construction}
\begin{algorithm}[h]
\caption{HMSG Construction}
\label{alg:graph}
\begin{algorithmic}[1]
\Require Floor–room–object layout, views with poses
\Ensure Hierarchical multi-modal scene graph $G$
\State Initialize empty graph $G$
\For{each floor $f$}
    \State Add node($f$) to $G$
    \For{each room $r$ in $f$}
        \State Add node($r$) and edge($f, r$)
        \For{each image view $v$ in $r$}
            \State Add node($v$) and edge($r, v$)
        \EndFor
        \For{each object $o$ in $r$}
            \State Add node($o$) and edge($r, o$)
            \State best\_view $\gets$ None, min\_depth $\gets \infty$
            \For{each view $v$ in $r$ where $o$ is visible}
                \State Add edge($v, o$)
                \If{mean\_depth($o, v$) $<$ min\_depth}
                    \State best\_view $\gets v$, min\_depth $\gets$ mean\_depth($o, v$)
                \EndIf
            \EndFor
            \State Assign best\_view to $o$
        \EndFor
    \EndFor
\EndFor
\State \Return $G$
\vspace{-2pt}
\end{algorithmic}
\end{algorithm}
Following HOVSG \cite{werby23hovsg}, we first construct an open-vocabulary map at the instance-level. Objects and views are associated with rooms based on geometric overlap in top-down maps, resulting in a scene graph enriched with both geometric and semantic features.

As shown in Algorithm \ref{alg:graph}, the Hierarchical Multi-modal Scene Graph (HMSG) takes a floor–room–object layout and posed image views as input, and outputs a directed graph $G$, where nodes represent semantic entities (floors, rooms, views, objects) and edges encode structural relations (e.g., room-in-floor, object-in-room). Floors and rooms are sequentially added with CLIP features; unlike HOVSG, FSR-VLN leverages GPT-4o to infer room names from image views. Each view is associated with CLIP embeddings, VLM-generated captions, and camera poses, while the edges connect visible objects to their corresponding views. Additionally, we compute the mean depth of each object across visible views and select the view with the minimum depth (closest appearance) as its representative. The resulting HMSG encodes hierarchical topology and multi-modal features-including semantic, geometric, and visibility information, providing the foundation for fast-to-slow reasoning to improve navigation performance.


\begin{figure*}[ht]
    \vspace{-6pt}    
    \centering
    \begin{overpic}[width=1.0\textwidth, keepaspectratio]{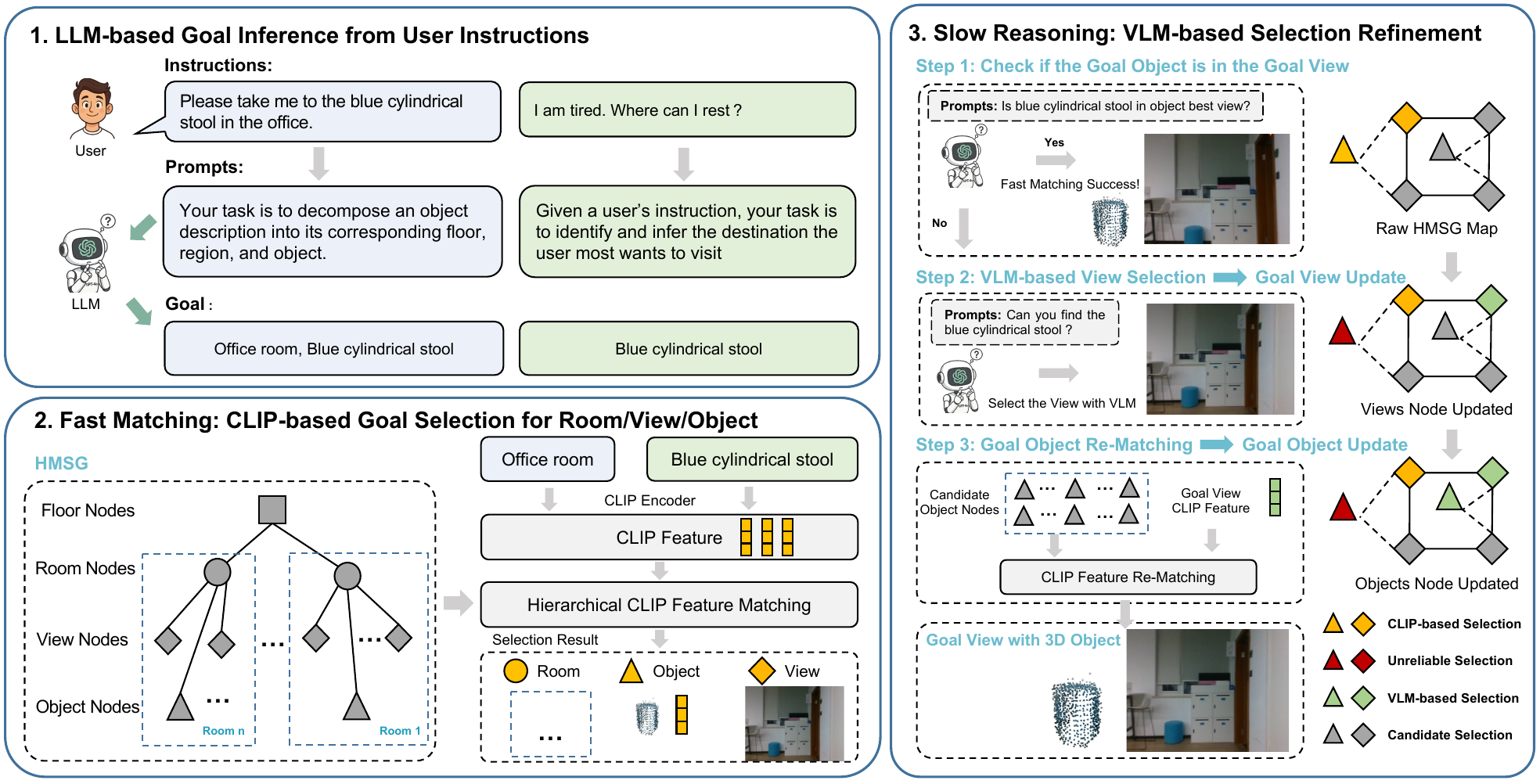}
    \end{overpic}
    \caption{The navigation reasoning follows a coarse-to-fine process: 1). LLM interprets user instructions into structured object/room queries; 2). CLIP-based fast matching, as intuition retrieves candidate goal rooms, views, and objects. 3). VLM-based slow reasoning refines the candidate results to ensure accurate goal view and object selection.} 
    \label{fig:nav_reasoning_pipeline}  
    \vspace{-6pt}
\end{figure*}

\subsection{LLM/VLM and Multi-modal feature-based Fast-to-slow Navigation Reasoning} 
\label{Coarse-to-fine navigation reasoning} 
As illustrated in Fig.~\ref{fig:nav_reasoning_pipeline}, our navigation reasoning comprises three steps: LLM-based user instruction understanding, Fast Matching, and Slow Reasoning.

\textbf{LLM-based user instruction understanding.} To interpret user instructions for object navigation, we introduce an LLM-based object query inference module that handles both spatially explicit and implicit natural language commands. For spatial instructions (e.g., “Take me to the blue cylindrical stool in the office”), the LLM acts as a hierarchical concept parser, decomposing the input into structured components such as floor, region, and object. The output maps directly to the nodes in the hierarchical scene graph, enabling precise localization.

For non-spatial instructions (e.g., “I’m tired, where can I find a blue cylindrical stool?” or “I’m thirsty”), the LLM functions as a goal inference agent, identifying the most relevant object or region based on user intent. The inferred object-level semantics are then resolved into spatial targets via the scene graph. This prompting framework supports generalization across diverse user expressions, both explicit and implicit, facilitating robust human-robot interaction in real-world environments.

\textbf{Fast Matching: Multi-modal feature-based Goal Room/View/Object Chosen.} After instruction understanding, if a room name is provided in the instruction, the system first matches the room and subsequently performs view and object matching within that room.
Using the image-view nodes in Fig.~\ref{fig:SceneGraph} and the LLM-interpreted text query, CLIP-based fast matching is performed between the query text and HMSG view-layer embeddings to identify the goal view. In parallel, object-level locations are determined by matching CLIP features between the query text and object embeddings, with the object exhibiting the highest similarity considered the potential target.
Although fast grounding of the goal view and object instances, the navigation is limited by the matching ability of the CLIP model, which may provide the wrong object. Thus, based on CLIP-based fast matching, VLM-based slow reasoning is further introduced to obtain a more precise goal view and object.

\textbf{Slow Reasoning: LLM/VLM-based View/Object Chosen Refinement.} After fast matching, the goal view, goal object, and other candidate views are obtained. Using the reasoning capabilities of VLM, we employ GPT-4o to verify whether the object is present in the best view corresponding to the matched object from the fast-matching step.
Since objects in the HMSG map are guaranteed to appear within their corresponding best views, if GPT-4o determines that the interpreted object is not present in the best view corresponding to the matched object, the matched object is considered unreliable.
Although fast matching may fail, the correct goal view may still exist among the matched views or unmatched candidates. To address this, we reapply both LLM and VLM reasoning to identify the optimal goal image. Specifically, we first use the LLM to reason over textual descriptions of unmatched views in the HMSG, selecting the most semantically consistent image as view-1. Next, the fast-matched view is compared with view-1, and VLM inference is applied to determine the final optimal goal image. Once the goal image is finalized, the goal object is updated by traversing its object list in the HMSG and recalculating the CLIP similarities between the query text and each object.

\section{Experiment}

\subsection{Experiment Setup}

\textbf{Dataset.} 
To validate our system in the real world, we use a Unitree-G1 humanoid robot equipped with a calibrated Intel RealSense D455 RGBD camera and a Mid360 LiDAR (Figure \ref{fig:system_overview}) to collect LiDAR–camera data across long-range office environments with long corridors and multiple rooms, denoted as Room1, Room2, Room3, and Room4. We also evaluated our approach on eight scenes from the HM3D-SEM dataset \cite{10204970}.

\textbf{User Instructions.}
To evaluate the generalization of our system across diverse instructions, we follow the experimental setup of MobilityVLA \cite{xu2024mobility} and crowd-source 87 user instructions in four categories: Reasoning-Free (RF) with explicit targets, Reasoning-Required (RR) requiring implicit goal inference, Small Objects (SO), which are challenging due to limited visual footprints, and Spatial Target (ST) designed to test long-range memory using room-type information. The instructions span 23, 18, 15, and 14 object categories, respectively.

\textbf{Metrics.}
In this paper, we focus on evaluating whether the target has been successfully retrieved. Following the metric in \cite{xu2024mobility, xie2025osmag}, we report the success rate (SR) and retrieval success rate (RSR$_{\text{top-}n@k}$), where the RSR is defined as the percentage of queries for which at least one of the top-$n$ predictions ($n \in {1, 5}$) lies within $k$ meters (Euclidean distance) of the ground truth. We evaluate performance with $k \in {1,2, 3, 4, 5},\text{m}$ to account for positional variance introduced by 2D image-derived nodes and 3D point clouds. Notably, when $n=1$ and $k=1$, SR is equivalent to RSR.

\textbf{Baselines.}
We evaluated FSR-VLN against several SOTA ObjNav methods using different map representations: CLIP-based 3D voxel maps (OK-Robot \cite{Liu_2024}), CLIP-based 3D scene graphs (HOVSG \cite{werby23hovsg}), and image-based topological graphs (MobilityVLA \cite{xu2024mobility}). OK-Robot performs object retrieval using CLIP-based OWL-ViT \cite{minderer2022simpleopenvocabularyobjectdetection} features. HOVSG retrieves floors, rooms, and objects based on CLIP features, with its 3D scene graph structure supporting long-range retrieval. MobilityVLA retrieves goal images from all views; as it is not open-sourced, we implemented a version using GPT-4o, first selecting the top 50 candidate frames via CLIP similarity before inference due to GPT-4o’s context limitations.
OK-Robot and MobilityVLA lack room-level information and cannot interpret Spatial Target (ST) instructions referencing a specific room. For fair comparison, we consider only the object query derived from LLM-based instruction understanding for these methods and HOVSG. Retrieval is also counted as successful if objects of the same type from other rooms are returned for ST instructions.
In contrast, FSR-VLN leverages room-level spatial information, and retrieval is considered successful only when the target object is in the specified room. By integrating the HMSG map with navigation reasoning, FSR-VLN jointly predicts the goal view and object, and a retrieval is successful only if both are correctly identified.
 
In the HM3D-SEM dataset, as osmAG-LLM, we select HOV-SG and osmAG-LLM as our baseline because they share fundamental design principles with our approach\cite{xie2025osmag}.

\subsection{Benchmark Results}
\textbf{Quantitative Analysis.}


\begin{table*}[ht]
\vspace{-5pt}
\scriptsize
\setlength{\tabcolsep}{5pt} 
\centering
\caption{Object Retrieval Comparison with SR, RSR@Top1, and RSR@Top5 metrics. 
The metrics are visualized \colorbox{green1}{green} (best) to \colorbox{red2}{red} (worst).}
\label{tab:results_color}
\begin{tabular}{lccccccccccccc}
\toprule
\multirow{2}{*}{\textbf{Method}}& \multirow{2}{*}{\textbf{Map Representation\&Navigation}} & \multirow{2}{*}{\textbf{Time}} & \multirow{2}{*}{\textbf{SR}} & \multicolumn{5}{c}{\textbf{RSR@Top1}} & \multicolumn{5}{c}{\textbf{RSR@Top5}} \\
\cmidrule(lr){5-9} \cmidrule(lr){10-14}
& & & & 1m & 2m & 3m & 4m & 5m & 1m & 2m & 3m & 4m & 5m \\
\midrule

OK-Robot\cite{jin2025openfusionopenvocabularyrealtimescene} & CLIP based 3D Voxel Map & 0.2s
& \getcolor{0.609}0.609 & \getcolor{0.609}0.609 & \getcolor{0.609}0.609 & \getcolor{0.609}0.609 & \getcolor{0.609}0.609 & \getcolor{0.609}0.609 
& \getcolor{0.632}0.632 & \getcolor{0.632}0.632 & \getcolor{0.632}0.632 & \getcolor{0.632}0.632 & \getcolor{0.632}0.632 \\
\hline
HOVSG\cite{werby23hovsg} & CLIP based 3D Scene Graph & 0.2s
& \getcolor{0.517}0.517 & \getcolor{0.517}0.517 & \getcolor{0.573}0.573 & \getcolor{0.586}0.586 & \getcolor{0.596}0.596 & \getcolor{0.596}0.596  
& \getcolor{0.770}0.770 & \getcolor{0.816}0.816 & \getcolor{0.828}0.828 & \getcolor{0.828}0.828 & \getcolor{0.828}0.828 \\
\hline
MobilityVLA\cite{xu2024mobility} & Image based Topological Graph + VLM & 30s
& \getcolor{0.345}0.345 & \getcolor{0.345}0.345 & \getcolor{0.598}0.598 & \getcolor{0.759}0.759 & \getcolor{0.805}0.805 & \getcolor{0.954}0.954 
& - & - & - & - & - \\
\hline
FSR-VLN (Ours) & Multimodal 3D Scene Graph + VLM & 5.5s
& \getcolor{0.920}0.920 & \getcolor{0.920}0.920 & \getcolor{0.943}0.943 & \getcolor{0.943}0.943 & \getcolor{0.966}0.966 & \getcolor{0.966}0.966 
& - & - & - & - & -  \\
\bottomrule
\end{tabular}
\vspace{-5pt}
\end{table*}

The performance of FSR-VLN compared to other methods on real-world datasets is summarized in Table \ref{tab:results_color}, reporting the average success rate (SR) over 87 instructions across four evaluation sets. FSR-VLN achieves the highest SR of 92\% (80/87), substantially outperforming baselines: MobilityVLA: 34.5\% (30/87), OK-Robot: 60.9\% (53/87), HOVSG: 51.7\% (45/87). This corresponds to relative improvements of 167\%, 51\%, and 77\%, respectively, demonstrating the effectiveness of our approach. 
A similar trend is observed for RSR@Top1: FSR-VLN consistently achieves the best performance across distance thresholds, reaching 96.6\% (84/87) at 4-5 meters, indicating robust long-range retrieval. Importantly, retrieval in FSR-VLN is considered successful only if the target object is located within the specified room; even under this stricter criterion, it achieves the highest SR and RSR, highlighting the advantages of HMSG-based reasoning.

MobilityVLA, constrained by semantic–geometric mismatches \cite{chen2025astra} due to weak 3D spatial cues, exhibits the lowest short-range RSR$_{\text{top-}1@k}$ ($k=1,2$m). However, at larger distance tolerances (3–5m), it ranks second to FSR-VLN, reflecting its strong image-sequence reasoning capabilities.

OK-Robot, constrained by OWL-ViT-based retrieval, achieves an average SR of 61\%, while HOVSG, similarly limited by CLIP-like models, reaches 52\%. Both methods lack spatial verification of matched objects, often resulting in navigation failures.
In contrast, FSR-VLN leverages the HMSG representation, which encodes geometric, semantic, and topological relations across floors, rooms, views, and objects. Built on this representation, the VLM-based FSR mechanism first performs fast feature matching to retrieve candidate views and objects, followed by VLM reasoning for verification. When mismatches are detected, slow reasoning refines the final selection, thereby enhancing overall navigation performance.
\begin{table}[ht]
\vspace{-5pt}
\scriptsize
\setlength{\tabcolsep}{6pt} 
\centering
\caption{Object Retrieval Comparison with RSR@Top1 and RSR@Top5  on HM3DSEM. The metrics are visualized \colorbox{green1}{green} (best) to \colorbox{red2}{red} (worst).}
\label{tab:results_color2}
\begin{tabular}{lccccccccc}
\toprule
\multirow{2}{*}{\textbf{Method}} & \multirow{2}{*}{\textbf{Time}} & \multicolumn{3}{c}{\textbf{RSR@Top1}} & \multicolumn{3}{c}{\textbf{RSR@Top5}} \\
\cmidrule(lr){3-5} \cmidrule(lr){6-8}
& & 1m & 2m & 3m & 1m & 2m & 3m \\
\midrule

HOVSG\cite{werby23hovsg} & 0.2s
& \getcolor{0.52}0.52 & \getcolor{0.64}0.64 & \getcolor{0.70}0.70
& \getcolor{0.76}0.76 & \getcolor{0.82}0.82 & \getcolor{0.88}0.88 \\
\hline
osmAG-LLM\cite{10204970} & 3.0s
& \getcolor{0.28}0.28 & \getcolor{0.50}0.50 & \getcolor{0.69}0.69
& \getcolor{0.47}0.47 & \getcolor{0.83}0.83 & \getcolor{0.90}0.90 \\
\hline
FSR-VLN (Ours) & 5.5s
& \getcolor{0.87}0.87 & \getcolor{0.88}0.88 & \getcolor{0.88}0.88
& \getcolor{0.92}0.92 & \getcolor{0.92}0.92 & \getcolor{0.92}0.92\\
\bottomrule
\end{tabular}
\vspace{-5pt}
\end{table}

As illustrated in Table \ref{tab:results_color2}, the Top-1 retrieval success rate of osmAG-LLM is significantly lower than that of HOVSG. The main reason is that osmAG-LLM loses the hierarchically rich visual open-vocabulary feature representation present in HOVSG and instead relies solely on textual recognition results. Although the osmAG-LLM semantic map is also hierarchical, it only preserves XML-level textual information without retaining visual CLIP features. In contrast, our method not only preserves CLIP-based visual embeddings but also further interacts with the original image information through VLM, leading to a notable improvement in retrieval success.

\begin{figure}[ht]
    \vspace{-6pt}    
    \centering
    \begin{overpic}[width=0.5\textwidth, keepaspectratio]{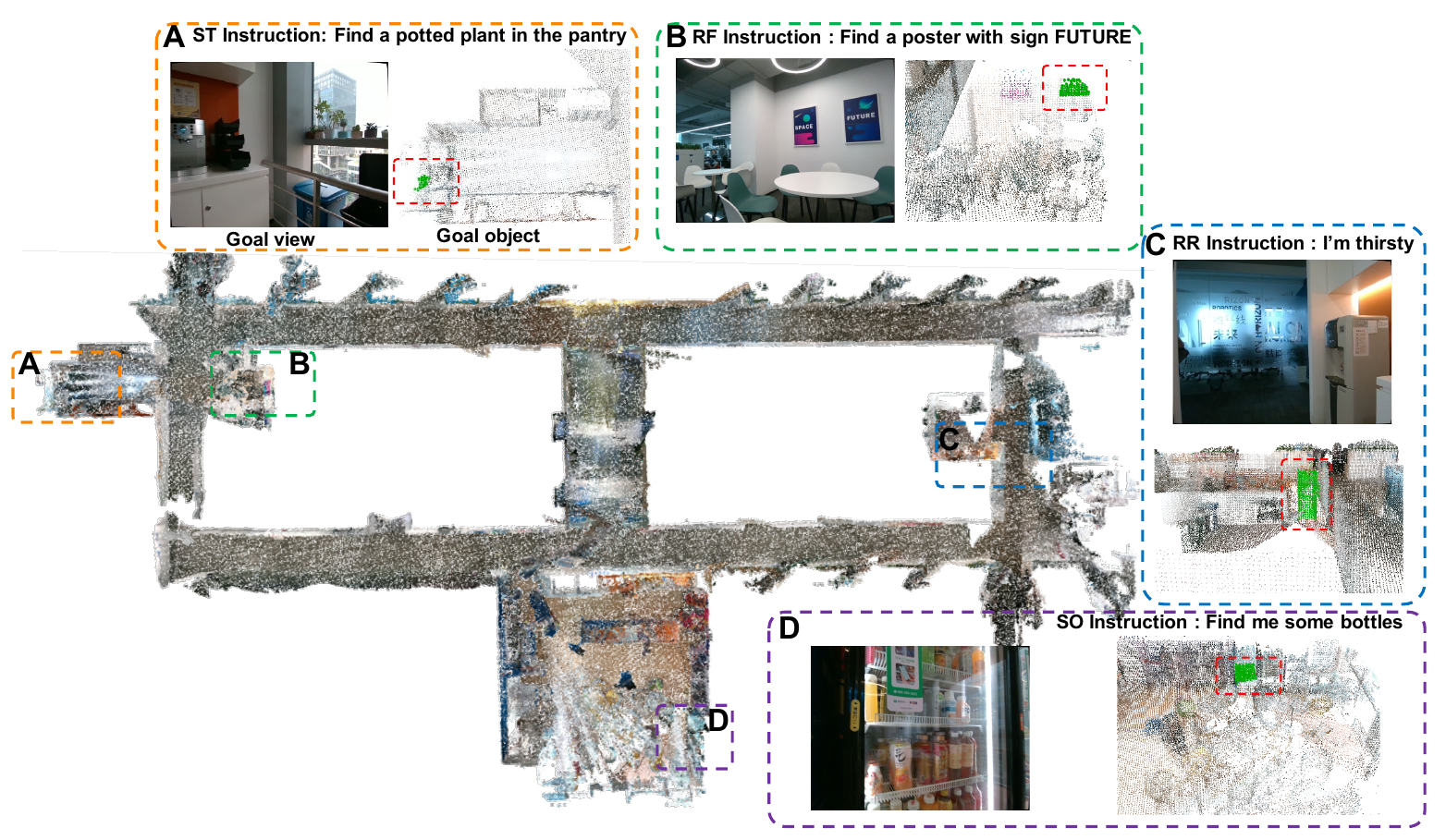}
    \end{overpic}
    \caption{The goal view and object retrieval results of FSR-VLN for four different instructions (Reasoning-Free, Reasoning-Required, Small Object, and Spatial Target) in Room4 (40mx20m).} 
    \label{fig:retrival_vis}  
    \vspace{-3pt}
\end{figure}

\textbf{Qualitative analysis.}
Fig.~\ref{fig:retrival_vis} illustrates the goal view and object retrieval results of FSR-VLN for different instructions in Room4. FSR-VLN successfully retrieves the goal view and object across four instruction types in long-range indoor environments.

\textbf{Run-Time Analysis.}
MobilityVLA is the slowest, requiring approximately 30 s to reason over image view sequences. In contrast, OK-Robot and HOVSG compute text–object similarity using OWL-ViT and CLIP in only 0.2 s. FSR-VLN achieves an average response time of only 1.5 s when relying solely on goal inference and fast matching, and 5.5 s when incorporating slow reasoning. By invoking slow reasoning only when fast matching fails, and applying VLM refinement to candidate views rather than entire sequences, FSR-VLN reduces average response time by 82\% compared with MobilityVLA while improving success rates. Within the integrated navigation system, FSR-VLN runs in parallel with user interactions, ensuring seamless and responsive operation.


\begin{table}[t]  
\vspace{-3pt}
\scriptsize
\setlength{\tabcolsep}{4pt} 
\centering
\caption{Comparison of RSR@Top1 Performance Across Different FSR-VLN Settings.
The highest value is highlighted in \textbf{bold}.}
\label{tab:results_ablations}
\begin{tabular}{lcccccc}
\toprule
\multirow{2}{*}{\textbf{Method}} & \multirow{2}{*}{\textbf{Time}} & \multicolumn{5}{c}{\textbf{RSR@Top1}} \\
\cmidrule(lr){3-7}
& & 1m & 2m & 3m & 4m & 5m \\
\midrule
Ours (wo ST / wo NR) & 1.5s
& 0.724 & 0.747 & 0.770 & 0.805 & 0.805 \\
\hline
Ours (w ST / wo NR) & 1.5s
& 0.816 & 0.862 & 0.885 & 0.908 & 0.908 \\
\hline
Ours (w ST / w NR) & 5.5s
& \textbf{0.920} & \textbf{0.943} & \textbf{0.943} & \textbf{0.966} & \textbf{0.966} \\
\bottomrule
\end{tabular}
\vspace{-6pt}
\end{table}

\subsection{Ablations Analysis}
To further validate the effectiveness of HMSG representation and FSR, we examine the impact of spatial instructions and the FSR process on FSR-VLN performance, as summarized in Table \ref{tab:results_ablations}. Without \textbf{Navigation Reasoning (NR)} and \textbf{Spatial Target (ST)} instructions guide long-range navigation is guided by restricting object search to the target room, reducing global matching errors, and improving RSR. In long-range environments, this room-level guidance is particularly critical; when combined with the hierarchical scene graph, the search is further confined to the designated room, enhancing navigation success. With NR, extended VLM reasoning allows FSR to verify the correctness of fast matching, and VLM-based selection refinement further increases RSR, demonstrating the effectiveness of the approach.

\section{Conclusion And Future Work}

We present FSR-VLN, a humanoid robotics VLN system that integrates a Hierarchical Multi-modal Scene Graph (HMSG) with Fast-to-Slow Navigation Reasoning (FSR). The HMSG encodes geometric, semantic, and topological relationships across floors, rooms, views, and objects, supporting FSR to fast retrieve candidate goals and refine them for more robust navigation. Experiments on real-world datasets demonstrate that FSR-VLN outperforms SOTA baselines in success rate, even under strict spatial constraints. These results highlight the effectiveness of HMSG and multi-stage FSR for robust navigation. Nevertheless, FSR-VLN has certain limitations. The construction of HMSG is time-consuming, making the approach unsuitable for real-time mapping, and it assumes static environments, limiting its applicability in dynamic settings. Our future work will address three key directions: enhancing the efficiency of scene graph construction, extending the system's robustness to dynamic environments, and integrating exploratory navigation capabilities to handle novel or ambiguous scenarios.



\addtolength{\textheight}{0cm}   



\bibliographystyle{IEEEtran}
\bibliography{References}
\end{document}